\documentclass[10pt,twocolumn,letterpaper]{article}

\usepackage{iccv}
\usepackage{times}
\usepackage{epsfig}
\usepackage{graphicx}
\usepackage{amsmath}
\usepackage{amssymb}
\usepackage{multirow}
\usepackage{bbm}
\usepackage[accsupp]{axessibility}

\usepackage[breaklinks=true,bookmarks=false]{hyperref}

\iccvfinalcopy 

\def\iccvPaperID{2067} 

\ificcvfinal\fi

\begin{document}

\title{Recurrent Mask Refinement for Few-Shot Medical Image Segmentation}

\author{ Hao Tang \qquad Xingwei Liu \qquad Shanlin Sun \qquad Xiangyi Yan \qquad Xiaohui Xie \\
Department of Computer Science\\
University of California, Irvine,  California, 92697\\
{\tt\small \{htang6, xingweil, shanlins, xiangyy4, xhx\}@uci.edu}
}

\maketitle
\ificcvfinal\thispagestyle{empty}\fi

\begin{abstract}
Although  having achieved great success in medical image segmentation, deep convolutional neural networks usually require a large dataset with manual annotations for training and are difficult to generalize to unseen classes.
Few-shot learning has the potential to address these challenges by learning new classes from only a few labeled examples. In this work, we propose a new framework for few-shot medical image segmentation based on prototypical networks. Our innovation lies in the design of two key modules: 1) a context relation encoder (CRE) that uses correlation to capture local relation features between foreground and background regions; and 2) a recurrent mask refinement module that repeatedly uses the CRE and a prototypical network to recapture the change of context relationship and refine the segmentation mask iteratively.
Experiments on two abdomen CT datasets and an abdomen MRI dataset show the proposed method obtains substantial improvement over the state-of-the-art methods by an average of 16.32\%, 8.45\% and 6.24\% in terms of DSC, respectively. Code is publicly available \footnote{\url{https://github.com/uci-cbcl/RP-Net}}.
\end{abstract}

\section{Introduction}
Medical image segmentation is a fundamental task in medical image analysis. It is used in many clinical applications, including disease diagnosis, treatment planning and treatment delivery. Segmentation of anatomical structures or lesions is usually done manually by experienced doctors, which is often tedious and labor-intensive. With the recent use of deep convolutional neural networks, automated segmentation tools using computer programs can achieve near human accuracy on multiple tasks with very short processing time. However, in order to achieve good performance, these systems are usually trained in a fully supervised fashion with large amounts of annotated data. Acquiring a dataset with abundant manual labels is often very expensive and time-consuming as it requires experts with many years' clinical experience. Moreover, the differences in image acquisition protocols among different medical equipment and institutes pose great challenges to the generalization ability of the learning based systems.

Few-shot learning has been proposed as one of the potential solutions to addressing these challenges in the low data regime \cite{snell2017prototypical,sung2018learning,vinyals2016matching,fei2006one,lake2011one}.
The main few-shot image segmentation approach forms the problem as meta learning \cite{finn2017model,finn2018probabilistic,hospedales2020meta} and uses supervised learning to train few-shot learning models. A few-shot learning model is trained to extract class-specific features from the set of support images with annotations, and then perform segmentation on the query images by using distilled knowledge from the support images. During test time, by extracting features from a set of new support images (unseen classes), the model is able to segment novel classes. Many few-shot learning methods have been proposed and achieved great performance on natural image segmentation tasks \cite{rakelly2018conditional,shaban2017one,dong2018few,Siam_2019_ICCV,wang2019panet,zhang2020sg,zhang2019pyramid,yan2019dual,hu2019attention}. However, applying few-shot learning models for medical image segmentation is still in early stages \cite{ouyang2019data,roy2020squeeze}.


Few-shot segmentation in medical images is different than that in natural images. Many approaches are based on prototypical networks \cite{snell2017prototypical}, and often apply masked average pooling \cite{dong2018few,wang2019panet,zhang2020sg} to extract class prototypes from feature maps within the foreground mask. This step usually assumes the masked region contains sufficient features to distinguish different classes, especially foreground and background. However, this may not always be true in medical images. Distinct local appearances and context information are more critical in determining the boundary for foreground and background. A clear boundary to separate regions of interest from the background is of critical importance in medical image segmentation. Moreover, the background is usually large and
spatially inhomogeneous while the foreground is small and
homogeneous \cite{ouyang2020self}, and there exists the abundance of tissues that share very similar appearance to each other, all of which add ambiguity to define the foreground and background regions. To address this issue, we encourage the network to explicitly model the context relationship between foreground and background pixels, especially pixels around the boundary.

\begin{figure*}
\begin{center}
\fbox{\includegraphics[width=0.9\linewidth]{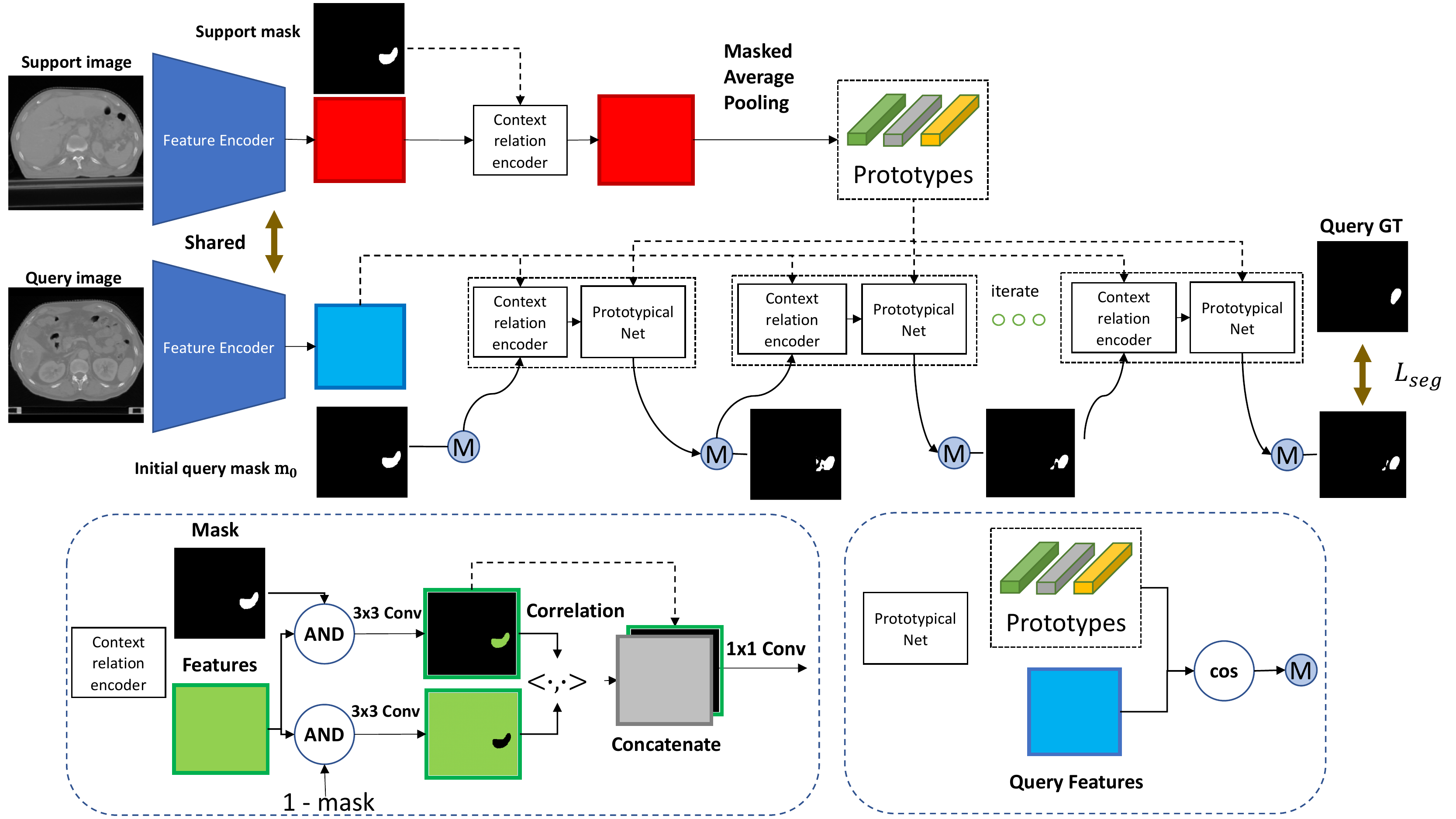}}
 
\end{center}
   \caption{RP-Net consists of three main components: (1) A feature encoder that extracts features from both support and query images; (2) A context relation encoder (CRE) that use correlation to enhance the local context relationship features; (3) A recurrent mask refinement module that iteratively uses CRE and a prototypical network to recaptures the change of local context features and refines the mask.}
\label{fig:model}
\end{figure*}

In this work, we introduce a new network framework for few shot medical image segmentation using prototypical network (RP-Net: \textbf{R}ecurrent \textbf{P}rototypical \textbf{Net}works). First, we propose a context relation encoder (CRE) on top of the extracted features, to explicitly model the relation between foreground and background feature maps. The relationships between foreground and background regions are more important in defining the boundary of the regions of interest in medical image segmentation. To force the model to distill and utilize the local context relation information, CRE uses correlation to capture the differences in the foreground and background regions. Pixel features are augmented with the context relation features. The explicit extraction of the context relationship poses a strong constraint to the features the model would learn and forces it to focus on the boundary of the region of interest. A prototypical network is followed to produce predicted masks using these augmented features.

Second, we propose a recurrent mask refinement module that iteratively refines the segmentation using CRE and prototypical networks. This design draws inspiration from recent works \cite{teed2020raft,peng2020deep,hur2019iterative} that employ iterative refinement. More importantly, the prediction mask modifies the mask in the previous step, which results in updated local context relationship. The recurrent module serves the purpose to recapture the updated context relationship and recompute its context relationship based on new prediction. Starting from the segmentation mask from the previous step, the model uses the refined prediction mask in the previous step to compute new context features using CRE, and then feeds it to the same prototypical network. The weights of the module are shared among multiple iterations so it is fully recurrent. This recurrent module facilitates the learning and forces the model to learn to gradually refine the segmentation. 

Our contributions are summarized as:

- A context relation encoder (CRE) that uses correlation between foreground and background to enhance context relationship features around the object boundary. 

- A new framework for few-shot medical image segmentation that iteratively refines the prediction mask through a recurrent module that uses CRE and prototypical networks.

- We conducted experiments on two abdomen CT datasets and one abdomen MRI dataset. Experiments show that the proposed framework outperforms the SOTA few-shot framework for medical image segmentation by an average of 16.32\% on ABD-110 dataset \cite{tang2021spatial}, 8.45\% on MICCAI15 Multi-Atlas Abdomen Labeling challenge dataset \cite{landman2015miccai} and 6.24\% on ISBI 2019 Combined Healthy Abdominal Organ Segmentation Challenge \cite{kavur2021chaos} in terms of DSC.

\section{Related work}

\subsection{Medical image segmentation}
In recent years, deep learning has brought significant progress to the field of medical image analysis \cite{shen2017deep}, such as computer-aided diagnosis \cite{setio2017validation,tang2018automated,tang2019end,tang2019nodulenet}, image registration \cite{balakrishnan2019voxelmorph,balakrishnan2018unsupervised,heinrich2019closing}, reconstruction \cite{you2018structurally,Yancy9175642,you2019ct}, and etc. In terms of medical image segmentation, the development of the deep convolutional neural networks has lead to various successful applications, including segmentation of tissue \cite{sun20193d,vu2019methods,nirschl2017deep}, anatomical structures \cite{tang2019clinically,chen2018voxresnet,tong2018fully,zhou2019prior,gibson2018automatic,dolz2018hyperdense,sun2020attentionanatomy,chen2021deep,ma2018multi,tang2019automatic} and lesions \cite{havaei2017brain,zhao2018deep,vorontsov2018liver,li2018h,seo2019modified,xu2019lstm}. One of the most famous and widely used network architecture is U-Net \cite{ronneberger2015u}. U-Net uses lateral connection to fuse features from encoders and decoders. Many its variants were proposed, with different focus on their designs. V-Net \cite{milletari2016v} extends the use of U-Net to 3D volume data. Attention U-Net \cite{oktay2018attention} proposes to use gated mechanism to filter features. nnUNet \cite{isensee2021nnu} combines different U-Net like network architectures and automatically configure the optimal setting for different tasks, which is the best out of box U-Net. These SOTA methods require abundant manual annotations for their specific tasks to achieve good performance. They are designed to fully utilize the power of annotated dataset, and is limited when segmenting novel classes.

\subsection{Few-shot learning}
Few-shot learning can be categorized into three main focuses: data, model and algorithm \cite{wang2020generalizing}. One main stream of few-shot segmentation in natural image that focuses on the model is prototypical networks \cite{snell2017prototypical}. Prototypical network uses the idea of meta learning \cite{finn2017model,finn2018probabilistic,hospedales2020meta} and applies averaged mask pooling to pool class-specific features from the support set, which is called prototypes. Then, segmentation for the query image is done by computing the cosine distance with each class prototype
. PANet \cite{wang2019panet} further improves upon this idea by proposing a prototype alignment network to better utilize the support set, by also predicting on support images using query images as support set.

In few-shot medical image segmentation, most works focus on generating new training data to enlarge the training set given only a few labels \cite{zhao2019data,mondal2018few,ouyang2019data,yu2020foal}. However, this still requires retraining the model when a new class needs to be segmented. More recently, a few works focus on designing network architecture that does not require retraining the model. Squeeze and excite \cite{roy2020squeeze} first proposes a few-shot learning architecture specifically designed for medical image segmentation. They propose to use squeeze and excite modules to fuse information from support image on to query image to guide the segmentation arm. \cite{ouyang2020self} proposes local prototypes to enrich the representation of class prototypes and a self-supervised training strategy using super pixels. Likewise, we focus on few-shot medical image segmentation without retraining the model, and we propose a new framework that uses CRE and recurrent mask refinement module to better capture local feature  and shape differences around foreground object boundary.

\section{Method}
We first describe the formal definition of few-shot medical image segmentation. Next, we introduce the architecture of RP-Net, especially the context relation encoder (CRE) and recurrent mask refinement module. 

\subsection{Problem definition}
In few-shot medical image segmentation task, the model is trained using images and a set of semantic labels $C_{tr}$ drawn from a training dataset $D_{tr}$. During inference, the model segments a new set of semantic classes $C_{te}$ from test images $D_{te}$, given a few labeled examples of $C_{te}$. Note that $C_{tr} \cap C_{te}=\emptyset$. For example, the model is trained using semantic labels $C_{tr}=\{\mbox{liver, left and right kidney}\}$ and during testing time the model needs to segment new semantic classes $C_{te} = \{\mbox{spleen}\}$. Let N be the number of semantic classes in $C_{te}$, and K be the number of examples for each semantic class in $C_{te}$. The few-shot learning problem is also referred to as N-way K-shot learning. In medical image segmentation, most works usually consider 1-way 1-shot learning \cite{roy2020squeeze,ouyang2020self}. 

To achieve the goal of segmenting unseen classes in inference time, an episodic training strategy is used widely \cite{wang2019panet,ouyang2020self,roy2020squeeze}. To simulate the situation in testing time where only K examples for each class are provided, the episodic training schema randomly draws each training example in the form of a support and query data pair [($\mathbf{x}_s, \mathbf{y}_s$), ($\mathbf{x}_q, \mathbf{y}_q$)] from $D_{tr}$. The model is trained to distill knowledge about a semantic class from the support set ($\mathbf{x}_s, \mathbf{y}_s$) and then apply this knowledge to segment query set $\mathbf{x}_q$. In inference time, only the K support images $\mathbf{x}_s$ and their corresponding labels $\mathbf{y}_s$ are given, and the model performs segmentation on query images $\mathbf{x}_q$.

\subsection{Proposed method}
We now introduce RP-Net for few-shot learning in medical images. For the rest of this section, we consider a 1-way K-shot learning problem. The architecture of RP-Net is shown in Figure \ref{fig:model}. Our approach consists of three steps: 1) extracting image features, 2) enhancing context relation features using CRE, 3) iteratively applying CRE and prototypical network to refine the segmentation mask. All stages are differentiable and can be trained end-to-end.

\subsubsection{Feature extraction}
The input to the network is a set of K support images $\mathbf{x}_s \in \mathbb{R}^{H\times W \times 1}$ and a query image $\mathbf{x}_q \in \mathbb{R}^{H\times W\times 1}$, padded to the same height H and width W. The support and query images are first aligned globally using affine transformation, which is a common step in many medical image tasks. 

The model first uses the same feature encoder $f_\theta$ to extract support features $\mathbf{F}_s \in \mathbb{R}^{H'\times W'\times Z}$ and query features $\mathbf{F}_q \in \mathbb{R}^{H'\times W'\times Z}$ respectively. H$'$ and W$'$ are the height and width of the feature map, and Z is the number of feature channels. An adapted version of the U-Net backbone was used as the feature encoder $f_\theta$. Instead of upsampling the feature maps to the original resolution as implemented in the original U-Net, we remove the last two upsampling blocks in the U-Net to save GPU memory and computation. This results in the resolution of the support and query features being 1/4 of the image resolution ($H'=H/4, W'=W/4$).

\subsubsection{Context relation encoder (CRE)}
In medical image segmentation, the local context features are important to determine the boundary of foreground and background. To strengthen and emphasize these features, we propose the context relation encoder to enhance context features and force the model to focus on the shape and context of the region of interest rather than pixels themselves. 

CRE takes the extracted features $\mathbf{F}$ (we drop subscript $q$ and $s$ for convenience) and foreground mask $\mathbf{m}$ as input and outputs augmented features $\mathbf{F}_{cre}=f_{cre}(\mathbf{F}, \mathbf{m}) \in \mathbb{R}^{H'\times W'\times Z}$. $\mathbf{m}$ is the mask of the foreground class from the support image ($\mathbf{y}_s$), or the proposed foreground mask of a query image. Features of foreground and background are first extracted by masking $\mathbf{F}$ using the mask $\mathbf{m}$: $\mathbf{F}_f=\phi_{f}(\mathbf{F}\odot \mathbf{m})$ and $\mathbf{F}_b=\phi_{b}(\mathbf{F}\odot (1-\mathbf{m}))$. $\phi_f$ and $\phi_b$ denote $3\times 3$ convolution. Next, a correlation computation is applied to acquire the context relation features between foreground and background feature vectors at each spatial location $(x,y)$ of $\mathbf{F}_b$ and $(x-i,x-j)$ of $\mathbf{F}_f$ with offset $i$ and $j$:
\begin{equation}
\label{equation:cre_c}
\begin{gathered}
\mathbf{C}^{(x,y,i,j)}=\sum_z \mathbf{F}_f^{(x,y,z)} \mathbf{F}_b^{(x-i,x-j,z)}
\end{gathered}
\end{equation}
Instead of computing correlation between every pair of pixels on $\mathbf{F}_f$ and $\mathbf{F}_b$, we limit the maximum displacement $d$ for comparison at each location $(x,y)$. Given a maximum displacement $d$, we only compute correlation $C^{(x,y,i,j)}$ in a neighborhood of size $2d+1$ by limiting the range of $(i,j)$. As a result, the context relation feature $\mathbf{C}$ is of size $H'\times W' \times (2d+1)^2$. $\mathbf{C}^{(x,y)}$ effectively captures information of how a background pixel is related to foreground when it is close to the object boundary. Finally, we concatenate $\mathbf{C}$ and $\mathbf{F}_f$ along channel dimension and apply a $1\times 1$ convolution to fuse foreground features and context relation features to obtain $\mathbf{F}_{cre}$. $d$ is set to 5 based on empirical results (see Table \ref{table:ablation} for details).  

Compared to directly computing correlation between feature maps, separating feature map into foreground and background features is important. Correlation calculated this way is sparse and has only non-zero values around the boundary, which captures the shape of the object and clearly differentiate a pixel from the background. Correlation calculated between full feature maps is not able to achieve this because it does not have the sense of boundary of the region.



\subsubsection{Prototypical networks}




Following \cite{ouyang2020self,wang2019panet}, we use a relative simple method for calculating the prototypes, averaging feature vectors within the mask and across support images. Given the enhanced image features of support set $\mathbf{F}_{cre,s}$, we first compute the prototype of class $c$ via masked average pooling:
\begin{equation}
\label{equation:prototype}
\begin{gathered}
\mathbf{p}_c=\frac{1}{K}\sum_{k=1}^K\frac{\sum_{x,y}\mathbf{F}_{cre,s}^{(k,x,y)}\mathbf{y}_s^{(k,x,y,c)}}{\sum_{x,y}\mathbf{y}_{s}^{(k,x,y,c)}}
\end{gathered}
\end{equation}
where $(x,y)$ is the index of pixels on the feature map, $(x,y,c)$ indexes the spatial locations of the binary mask of class $c$ and $K$ is the number of support images.

Segmentation is done using a non-parametric metric learning method. Prototypical network calculates the distance between the query feature vector and the computed prototypes $P=\{\mathbf{p}_c|c\in C\}$. A softmax over the distances is applied to produce a probabilistic output over all classes. Formally, for each pixel at location $(x,y)$ of query feature map $\mathbf{F}_{cre,q}$, we have:
\begin{equation}
\label{equation:m_soft}
\begin{gathered}
\mathbf{m}_{soft} = \operatorname*{cos}(\mathbf{F}_{cre,q}, P), \mbox{and}\\
\operatorname*{cos}(\mathbf{F}_{cre,q}, P)^{(x,y,c)}=\frac{\mbox{exp}(-\alpha d(\mathbf{F}_{cre, q}^{(x,y)}, \mathbf{p}_c))}{\sum_{\mathbf{p}_j\in P}\mbox{exp}(-\alpha d(\mathbf{F}_{cre,q}^{(x,y)}, \mathbf{p}_j))}
\end{gathered}
\end{equation}
where the distance function $d$ is a commonly used cosine distance and $\alpha$ is a scaling factor for this distance function to work best with the softmax function. $\alpha$ is set to 20 \cite{wang2019panet}. The class prediction can be obtained by:
\begin{equation}
\label{equation:m}
\begin{gathered}
\mathbf{m}^{(x,y)}=\operatorname*{arg\,max}_c \mathbf{m}_{soft}^{(x,y,c)}
\end{gathered}
\end{equation}

\subsubsection{Recurrent mask refinement}
Since the mask $\mathbf{m}$ used to compute context relation features would change every time the network makes a prediction, we propose a recurrent mask refinement module to recapture this change and compute new context relation features based on the previous prediction.

The recurrent mask refinement module estimates a sequence of mask predictions \{$\mathbf{m}_1, \mathbf{m}_2,...,\mathbf{m}_n$\} from an initial mask which is the union of all support masks: $\mathbf{m}_0=\bigcup_{i=1}^{K}\mathbf{y}_s^i$.
At each iteration t, it produces a new segmentation mask $\mathbf{m}_{t}$ based on $\mathbf{m}_{t-1}$. 
The design of this architecture mimics the steps of an optimization algorithm. For this purpose, all the weights in the recurrent module are shared across multiple iterations. The model is trained to learn to modify the mask gradually so that the final output mask $\mathbf{m}_n$ converges to an optimum solution. Note that, in this work the $\mathbf{m}_0$ is initialized using the average of support masks since images are affine aligned, but it is also possible to better initialize $\mathbf{m}_0$ using other methods.

This recurrent mask refinement module takes in support features $\mathbf{F}_s$, query features $\mathbf{F}_q$ and the mask $\mathbf{m}_{t-1}$ in previous step, uses CRE to enhance query features, and applies prototypical network to output a segmentation mask $\mathbf{m}_t$. 

\begin{equation}
\label{equation:m_soft_softmax}
\begin{gathered}
\mathbf{m}_{soft,t} = \operatorname*{cos}(f_{cre}(\mathbf{F}_q, \mathbf{m}_{soft,t-1}), P)
\end{gathered}
\end{equation}

We apply 4 iterations of the recurrent mask refinement module during training to save memory and computation cost. In inference time, we apply 10 iterations. We show in Figure \ref{fig:iter_performance} the performance at each iteration during inference time and 10 iterations are sufficient to obtain a stable result. The final prediction is obtained by upsampling $\mathbf{m}_n$ to the same resolution of the $\mathbf{x}_q$ using bilinear interpolation.

\subsubsection{Loss function}
We supervise our network using dice loss and cross entropy between the final predicted mask $\mathbf{m}_{soft,n}$ and ground truth segmentation mask $\mathbf{y}_q$:

\begin{equation}
\label{equation:dice_loss}
\begin{gathered}
L_{seg} = \beta L_{dice} + L_{ce} \\
L_{dice} = 1 - \frac{2\sum_{i,j,c} \mathbf{m}_{soft,n}^{(i,j,c)} \mathbf{y}_q^{(i,j,c)}}{\sum_{i,j,c}\mathbf{m}_{soft,n}^{(i,j,c)}+\sum_{i,j,c}\mathbf{y}_q^{(i,j,c)}}\\
L_{ce} = -\frac{1}{HWC}\sum_{i,j,c} \mathbf{y}_q^{(i,j,c)} \log(\mathbf{m}_{soft,n}^{(i,j,c)})
\end{gathered}
\end{equation}

where $\beta$ is a constant controlling the strength of the two loss terms and is set to 1. Note that the use of the sum of dice loss and cross entropy is widely used in medical image segmentation tasks, such as \cite{isensee2018nnu}.

\begin{table*}
\begin{center}
\begin{tabular}{|l|l|c|c|c|c|c|}
\hline
\textbf{Dataset} & \textbf{Method} & \textbf{Spleen} & \textbf{Kidney L} & \textbf{Kidney R} & \textbf{Liver} & \textbf{mean} \\
\hline
\hline


\multirow{7}{*}{ABD-110} & PANet-init \cite{wang2019panet}        & 30.95$\pm$1.09           & 19.24$\pm$0.37             & 17.64$\pm$0.71             & 49.91$\pm$0.34          & 29.43         \\
& PANet \cite{wang2019panet}          & 35.89$\pm$1.75           & 40.22$\pm$1.71             & 41.54$\pm$0.82             & 52.36$\pm$0.60          & 42.50         \\
& SE-Net \cite{roy2020squeeze} & 29.48$\pm$1.07           & 37.48$\pm$2.08             & 37.53$\pm$1.97             & 19.09$\pm$0.36          & 30.89         \\
& SSL-ALPNet  \cite{ouyang2020self}       & 64.90$\pm$1.62           & 61.58$\pm$2.53             & 64.05$\pm$2.27             & 71.83$\pm$1.81          & 65.59         \\
& Affine  & 50.42$\pm$0.91 & 53.04$\pm$1.57 & 52.025$\pm$2.17 & 66.99$\pm$1.20 & 55.62 \\
& RP-Net (Ours)                & \textbf{78.77$\pm$0.64}           & \textbf{81.89$\pm$1.45}             & \textbf{85.12$\pm$0.98}             & \textbf{81.88$\pm$0.63}          & \textbf{81.91}         \\
\cline{2-7}
& Fully supervised \cite{tang2021spatial}    & 95.9            & 95.7              & 95.7              & 96.4           & 95.92      \\
\hline
\hline

\multirow{6}{*}{ABD-30} & SE-Net \cite{roy2020squeeze}     & 0.23           & 32.83             & 14.34       & 0.27       & 11.91        \\
& PANet \cite{wang2019panet}      & 25.59                     & 32.34             & 17.37           & 38.42  & 29.42         \\
& SSL-ALPNet \cite{ouyang2020self} & 60.25                     & 63.34             & 54.82          & 73.65   & 63.02         \\
& Affine             & 48.99           & 43.44      & 45.67   & 68.93     & 51.75         \\
& RP-Net (Ours)       & \textbf{69.85$\pm$2.34}   & \textbf{70.48$\pm$2.55}        & \textbf{70.00$\pm$0.89}    & \textbf{79.62$\pm$0.91}     & \textbf{72.48}       \\
\cline{2-7}
& Fully supervised \cite{zhou2019prior} & 96.8  & 95.3 & 92.0 & 97.4 & 95.4 \\
\hline
\hline
\multirow{6}{*}{ABD-MR} & SE-Net \cite{roy2020squeeze}        & 51.80                     & 62.11             & 61.32        & 27.43     & 50.66         \\
& PANet \cite{wang2019panet} & 50.90                    & 53.45             & 38.64     & 42.26         & 46.33         \\
& SSL-ALPNet \cite{ouyang2020self}    & 67.02                 & 73.63             & 78.39   & 73.05           & 73.02         \\
& Affine                & 62.87                        & 64.70             & 69.10    & 65         & 65.41         \\

& RP-Net (Ours)                   & \textbf{76.35$\pm$0.66}                    & \textbf{81.40$\pm$2.10}             & \textbf{85.78$\pm$1.12}       & \textbf{73.51$\pm$1.55}      & \textbf{79.26}  \\    
\cline{2-7}
& Fully supervised \cite{isensee2018nnu} & - & - & - & - & 94.6 \\
\hline
\end{tabular}
\end{center}
\caption{DSC comparison with other methods on ABD-110, ABD-30 and ABD-MR (unit: \%).}
\label{table:abd-110}
\end{table*}

\begin{table*}
\begin{center}
\begin{tabular}{|l|l|c|c|c|c|c|}
\hline
\textbf{Experiment}                       & \textbf{Method}                                    & \textbf{Spleen} & \textbf{Kidney L} & \textbf{Kidney R} & \textbf{Liver} & \textbf{mean} \\
\hline
\hline
\multirow{6}{*}{Added components} 
                                    & Affine                                             & 50.42         & 53.04             & 52.025            & 66.99          & 55.62         \\
                                    & Affine + Grabcut                                   & 57.93           & 64.17             & 64.25             & 65.27          & 62.91         \\
                                    & Affine + Concat                                    & 56.41         & 52.39             & 54.99             & 70.87          & 58.66         \\
                                    & Affine + CRE                                       & 57.73           & 58.05             & 60.62           & 73.53          & 62.48         \\
                                    & Affine + Concat + Recurrent                        & 59.99           & 60.65             & 62.31             & 83.03          & 66.50         \\
                                    & \underline{Affine + CRE + Recurrent} & 78.77           & 81.89             & 85.12             & 81.88          & 81.91         \\
\hline
\multirow{3}{*}{Backbone}           & VGG16                                              & 73.57           & 67.49             & 56.81             & 72.04          & 67.48         \\
                                    & Res18                                              & 72.39           & 79.13             & 81.61             & 80.89          & 78.50         \\
                                    & \underline{U-Net}                    & 78.77           & 81.89             & 85.12             & 81.88          & 81.91         \\
\hline
\multirow{4}{*}{Correlation radius} & $d=0$ & 78.40 & 81.90 & 82.12 & 83.89 & 81.58 \\

& $d=1$  & 80.03 & 81.87 & 82.09 & 82.1  & 81.52 \\
                                    & $d=3$ & 79.12 & 81.79 & 83.41 & 81.32 & 81.41 \\
                                    & \underline{$d=5$}                                                & 78.77           & 81.89             & 85.12             & 81.88          & 81.91         \\
                                    & $d=7$                                                & 77.56           & 80.25             & 81.77             & 80.22          & 79.95     \\
\hline
\multirow{4}{*}{Initialization} & Affine          & 50.42 & 53.04 & 52.02 & 66.99 & 55.62 \\
                                & Demons          & 63.60                     & 63.89                     & 61.89                      & 73.59                     & 65.74                     \\
                                & \underline{RP-Net (Affine)} & 78.77                     & 81.89                     & 85.12                      & 81.88                     & 81.91                     \\
                                & RP-Net (Demons) & 80.31                     & 83.55                     & 85.01                      & 82.86                     & 82.93   \\
\hline
\end{tabular}
\end{center}
\caption{Ablation study on ABD-110 (unit: \%). Underlined is the final configuration used in RP-Net.}
\label{table:ablation}
\end{table*}

\section{Experiment}
\subsection{Setup}
 \textbf{Dataset} We conducted experiments using two abdomen CT datasets and one MRI dataset:

- ABD-110 is an abdomen dataset from \cite{tang2021spatial} that contains 110 3D CT images from patients with various abdomen tumors and these CT scans were taken during the treatment planning stage. 

- ABD-30 is an abdomen dataset from the MICCAI 2015 Multi-Atlas Abdomen Labeling challenge \cite{landman2015miccai}. It contains 30 3D abdominal CT scans (ABD-30) from patients with various pathologies and has variations in intensity distributions between scans.

- ABD-MR is a MRI dataset from ISBI 2019 Combined Healthy Abdominal Organ Segmentation Challenge \cite{kavur2021chaos}. It contains 20 3D T2-SPIR MRI scans.

We perform the same 5-fold cross validation and consider only 1-way 1-shot learning, following the same protocol as previous work setting 2 \cite{ouyang2020self}. Liver, spleen and left and right kidney are used as semantic classes. Within each fold, one organ is considered as unseen semantic class for testing while the rest are used for training. Moreover, to reduce the variance by choosing only one support image during inference, following \cite{wang2019panet}, for each query image in the test set we randomly sample one support image from the test set, repeat this process for 5 times and the final result is obtained by averaging the 5 runs. 

\textbf{Evaluation metric} We use the same evaluation metric Sørensen–Dice coefficient (DSC) as in previous work \cite{ouyang2020self,roy2020squeeze}. DSC measures the overlap of the prediction mask $\mathbf{m}$ and ground truth mask $\mathbf{g}$, and is defined as:

\begin{equation}
\label{equation:dsc_score}
\begin{gathered}
\mbox{DSC}(\mathbf{m}, \mathbf{g})=\frac{2|\mathbf{m} \cap \mathbf{g}|}{|\mathbf{m}|+|\mathbf{g}|}
\end{gathered}
\end{equation}

\textbf{Implmentation details} All images are resampled to have the same $xy$-plane spacing of 1.25mm $\times$ 1.25mm. For segmenting 3D volume data, we follow the same protocol used in \cite{ouyang2020self,roy2020squeeze} by dividing the support and query images into 12 chunks and segmenting all slices in the query chunk by using the center slice in the corresponding chunk of the support image. During training, a pair of support and query images and their labels are both cropped to have a fixed size of 256 $\times$ 256 around the image center. Support and query images are aligned online using affine transformation before feeding into the network. RP-Net is trained from scratch using Adam as optimizer with initial learning rate 0.0001 for 50 epochs and the learning rate is reduced by a factor of 10 every 20 epochs. We also add the alignment loss to train RP-Net as in \cite{wang2019panet}.

\subsection{Comparison with the state-of-the-art methods}
Table \ref{table:abd-110}
shows the performance comparison of RP-Net with previous work on ABD-110, ABD-30, ABD-MR respectively. PANet \cite{wang2019panet} is an extended version of the widely used prototypical network \cite{snell2017prototypical} designed for natural image segmentation. PANet-init means directly using the pretrained VGG16 feature extraction backbone without any finetuning on the few-shot setting. SE-Net \cite{roy2020squeeze} is the first specifically designed architecture for few-shot medical image segmentation.  SSL-ALPNet \cite{ouyang2020self} is the state-of-the-art few-shot medical image segmentation framework that uses self-supervised learning and prototypical networks. Affine is the result of the accuracy after globally aligning the support and query image using affine transformation, which we use as an initial mask. \cite{ouyang2020self} reported performance for PANet-init, PANet, SE-Net and SSL-ALPNet on ABD-30 and ABD-MR, so these numbers are directly quoted. We ran these algorithms using public available code to report their performance on ABD-110.

First, compared to PANet, RP-Net outperforms PANet by 39.49\%, 43.06\% and 21.75\% on the three datasets ABD-110, ABD-30 and ABD-MR respectively. Second, compared to SE-Net, RP-Net outperforms SE-Net by 51.02\%, 60.57\% and 27.42\% on ABD-110, ABD-30 and ABD-MR respectively. Third, compared to the state-of-the-art method SSL-ALPNet, RP-Net outperforms SSL-ALPNet by an average of 16.32\%, 9.46\% and 6.24\% on ABD-110, ABD-30 and ABD-MR respectively.

These experiments demonstrate our approach can achieve the SOTA accuracy on medical image datasets with different image modalities (CT and MRI). Also, we focus on designing a new framework for few-shot medical image segmentation, which outperforms other approaches of the same motivation, e.g. SE-Net by a large margin. Additional gain may be obtained by combining our method with the self-supervised training schema proposed in SSL-ALPNet. 

\subsection{Ablation study}

Ablation experiments are conducted using the ABD-110 dataset, because it has more data compared to the other two. Table \ref{table:ablation} shows the results for the following experiments.

\textbf{Effect of each component} To verify the contribution of the two added components - context relation encoder and recurrent module, we conducted experiments by adding one component at a time: 1) model trained and tested without the CRE. To make use of the support mask which is used in CRE, we concatenate the mask to the feature map from backbone and apply a $3\times 3$ convolution for a fair comparison (denoted as concat). 2) model trained without recurrent module. Note that if we remove both CRE and recurrent training, the model becomes the PANet \cite{wang2019panet}. Moreover, we compare with Grabcut \cite{rother2004grabcut} which is an unsupervised method that uses iterated Graphcut. Grabcut can be seen as an unsupervised version of our algorithm. 

First, we verify the effect of using CRE. Affine + Concat is a naive way of integrating support masks by concatenating it directly to the feature maps, which outperforms the Affine by 3.04\%. Affine + CRE implements the more sophisticated way of exploring local feature differences using CRE, which outperforms the Affine + Concat by 3.82\%. This shows the CRE better captures the local difference via the use of correlation. However, the performance improvement is still not significant and the reason is that the mask prediction is changed each time and it lacks a mechanism to recapture this change and recompute the new local differences. The recurrent mask refinement module serves this purpose and we discuss its effect in the next paragraph.

Second, we compare the performance of using the recurrent mask refinement module. Affine + Concat + Recurrent means we apply the recurrent module to the concatenated feature map, which performs 7.84\% better than not using the recurrent module (Affine + Concat). This shows that the recurrent training indeed helps the model to find the right mask prediction because the initial mask from support is a very rough estimation of the location of the region of interests. If we combine the two added components together (Affine + CRE + Recurrent), we can achieve a big improvement by 15.39\% compared to Affine + Concat + Recurrent. This demonstrates that the integration of recurrent module to recapture local changes in the CRE is very important and can greatly boost the performance.


Third, we compare with Grabcut. Our method is in some sense similar to Grabcut - we both use an iterative update to refine the segmentation mask. Grabcut outperforms the baseline Affine by 7.29\%, showing that iteratively refining a mask is indeed beneficial. RP-Net (Affine + CRE + Recurrent) outperforms Grabcut by 19\%. There are mainly three reasons for this large improvement. First, Grabcut only uses one image, thus only image intensity is used to separate foreground and background region. On the contrary, RP-Net uses the support images to extract knowledge about the relationship between the foreground and background region, and utilize this knowledge to guide the segmentation of the new image. Second, Grabcut only refines the mask in the probable foreground region which is a human defined boundary and lacks the flexibility to attend other areas in the image, as well as the ability to correct error in the sure foreground region. RP-Net does not have these constraints and can potentially use information from the whole image. Third, RP-Net uses training data to train the feature extractor, while Grabcut is not a learning-based method and only uses information directly derived from pixel intensity.


\textbf{Effect of feature extraction backbone} We also experimented with three different feature extraction backbones - VGG16 \cite{simonyan2014very}, Res18 \cite{he2016deep} and U-Net \cite{ronneberger2015u}. To make sure the output feature map is 1/4 of the original image resolution for a fair comparison, we only kept the first two downsampling operations in both VGG16 and Res18 backbones and the rest of the network architecture remained the same. As seen from Table \ref{table:ablation}, VGG16 backbone performs the worst among the three backbones, which is 8.03\% lower than Res18. U-Net backbone outperforms Res18 backbone by an average of 2.32\% which is mainly because of the lateral connection in U-Net that fuses both low-level and high-level features. 
This demonstrates that RP-Net is compatible with different backbones, and backbones that perform better on medical image segmentation task, such as U-Net, would result in similar gain when combined with RP-Net.

\textbf{Effect of correlation radius} We conducted experiments with different radius $d=0,1,3,5,7$ in the correlation layer, which controls how many neighbouring pixels are included when computing correlation. $d=0$ means the correlation computation is carried out only at a single point. Note that even with $d=0$, the model is able to use features from the surrounding pixels because $\phi_f$ and $\phi_b$ are used to extract foreground and background specific features. Table \ref{table:ablation} shows
our approach is not very sensitive to the radius, and this is likely because RP-Net is designed to focus on a small region around the object boundary at a time, a larger context may not necessarily bring more benefits.

\begin{figure}
\begin{center}
\includegraphics[width=0.9\linewidth]{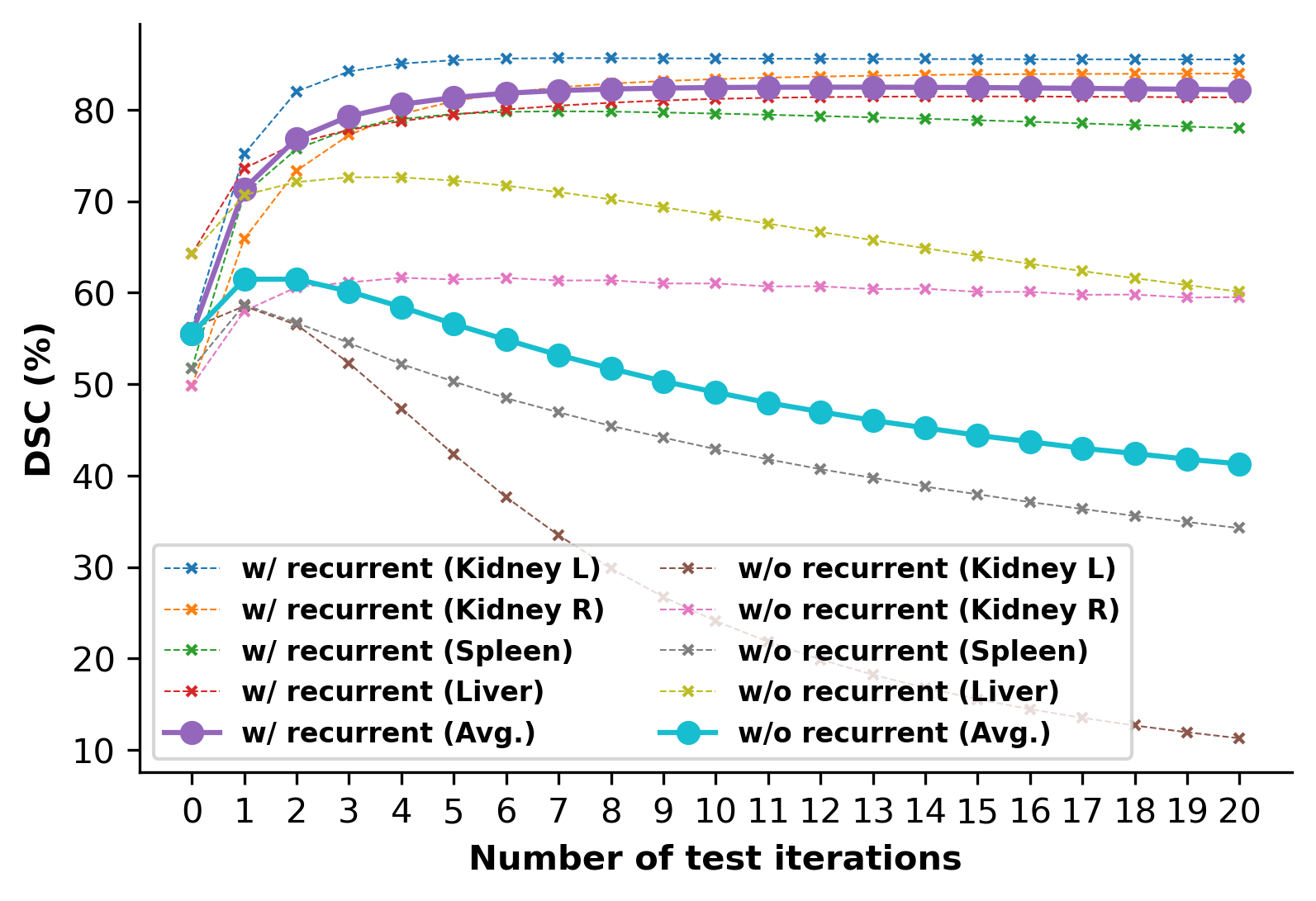}
 
\end{center}
   \caption{DSC at each refinement iteration. This figure shows the DSC performance of the proposed model per iteration. DSC of four organs and an average is shown for two models: one w/ recurrent training (purple) and one w/o recurrent training (cyan).}
\label{fig:iter_performance}
\end{figure}

\textbf{Effect of number of inference iterations} We show in Figure \ref{fig:iter_performance} the performance at each inference iteration from one fold in ABD-110. Although the model is trained using 4 iterations of recurrent module, we can apply more iterations during inference. As seen from this figure, a model without recurrent training diverges after the 1st iteration, while a model with recurrent training quickly converges and does not diverge after 20 epochs. It demonstrates that with the recurrent training, the model learns to gradually refine its prediction and converges to a stable solution. 

\textbf{Effect of initialization} Demons \cite{thirion1998image} is a medical image registration method that uses deformable registration, which performs 10.12\% better than a simple affine transformation. As shown in Figure \ref{table:ablation}, using a better initialization (Demons), RP-Net achieves a 1.02\% improvement. Although better initialization improves the result, the improvement is small compared to that of the initialization itself, and our network is less sensitive to the initial mask as long as it roughly locates the foreground region. For this reason, we only use initialization mask from Affine transformation for its simplicity. In many cases, a coarse map or a map derived through affine registration would suffice. Some recent registration methods (e.g., DEEDS \cite{heinrich2015multi} and its extensions) that can handle large anatomical variations, although missing details, can fit well to our method.

\begin{figure}
\begin{center}
\includegraphics[width=\linewidth]{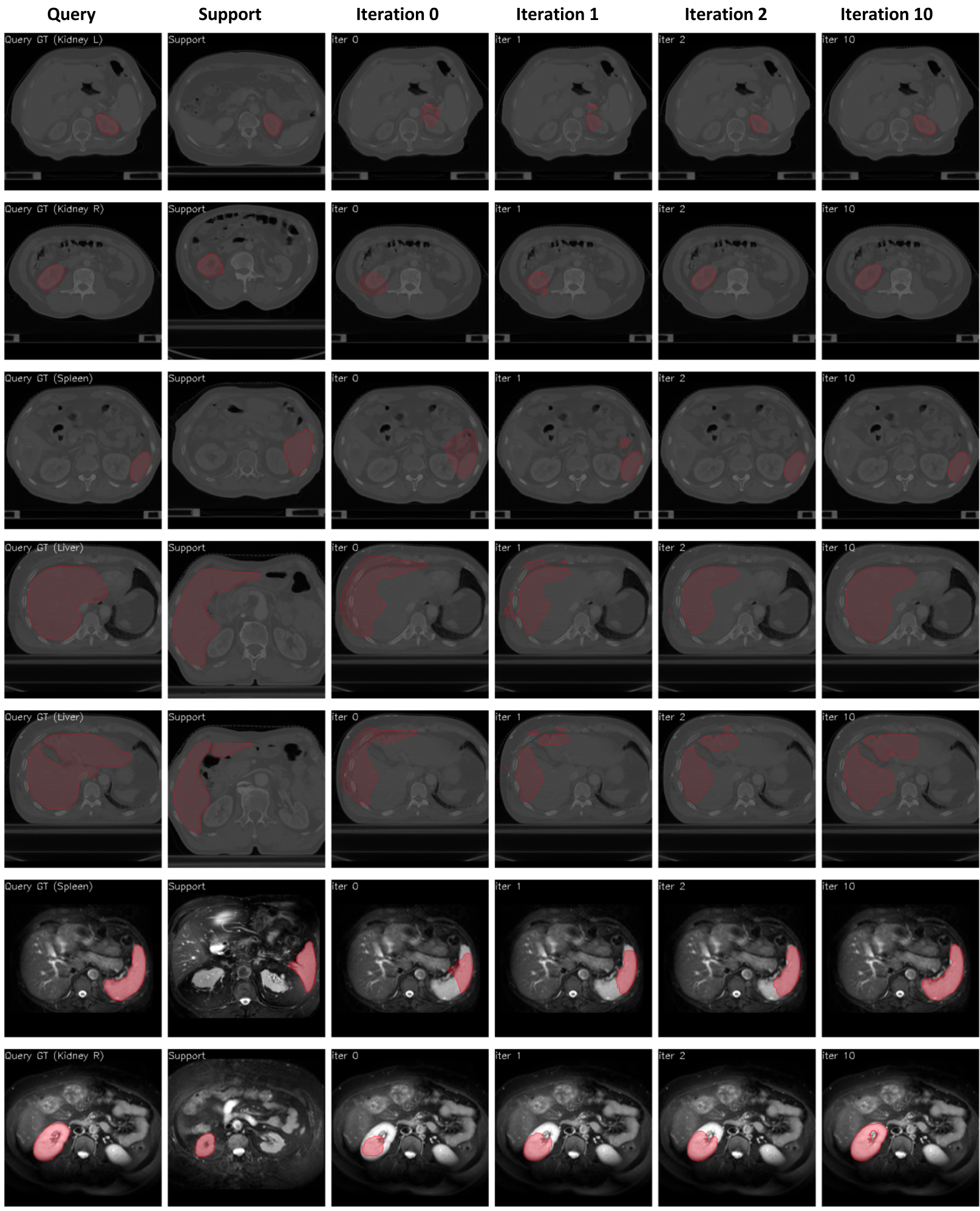}
 
\end{center}
   \caption{Examples of predication of RP-Net at different iterations. Each row represents one slice of the a test scan (row 1-5 are CT images, row 6-7 are MR images). 
}
\label{fig:good_examples}
\end{figure}

 

\subsection{Qualitative result}
We show in Figure \ref{fig:good_examples} how the segmentation mask converges to the optimum solution in multiple iterations. In general, we can observe that RP-Net refines the initial mask gradually, finds a better segmentation mask at each iteration, and finally converges to an optimum solution. 
RP-Net is able to learn to distill knowledge about the relation between the foreground and background from the support image, and apply it to segment query images by comparing local differences and modifying its prediction to conform to the shape and boundary. 
Moreover, RP-Net generates satisfying segmentation masks that have a clear boundary along the object boundary, demonstrating the successful design of the CRE and recurrent module. 


\section{Conclusion}
In this work, we present a new few-shot medical image segmentation framework that refines the segmentation mask iteratively using a context relation encoder and a recurrent module. The proposed model learns to incrementally refine the segmentation mask to better align the object boundary. Experiments on three organ segmentation datasets demonstrate that RP-Net outperforms the previous state-of-the-art approach by as much as 16\% in terms of DSC. 
Moreover, the proposed CRE and recurrent module are generic and can also be integrated into other types of network to enhance context relationship features.

{\bf Acknowledgement}
This work is partly supported by NSF grant IIS-1715017, a Simons Foundation grant (594598), and a hardware grant from NVIDIA. 

{\small
\bibliographystyle{ieee_fullname}
\bibliography{egbib}
}

\end{document}


\title{Recurrent Mask Refinement for Few-Shot Medical Image Segmentation}

\author{ Hao Tang \qquad Xingwei Liu \qquad Shanlin Sun \qquad Xiangyi Yan \qquad Xiaohui Xie \\
Department of Computer Science, University of California Irvine\\
University of California Irivne, Irvine, California, 92697\\
{\tt\small \{htang6, xingweil, shanlins, xiangyy4, xhx\}@uci.edu}}

\maketitle

\section{Supplementary \#1}

Regarding the selection of support and query images, we used randomly chosen pairs of support and query images, instead of one fixed support, in order to make sure that our results are not biased by a particular choice of support/query pairs in Table 1. Among the three datasets (ABD-110, ABD-30 and ABD-MR) we tested, the selection of support/query images on ABD-110 are exactly the same for all compared methods. We ensured the same sequence of support/query pairs (albeit random) were used by fixing the seed of the random number generator. Our selection of support/query images was different for ABD-30 and ABD-MR from \cite{ouyang2020self}, which used a fixed support image. To ensure the results are directly comparable, we reran our method following exactly the same selection of support as in \cite{ouyang2020self}. Our conclusion remains the same (Supplementary \Cref{table:abd-110}). Our result (labeled ``RP-Net (fixed support)") shows the proposed method still outperforms the SOTA method by a large margin.

\begin{table*}
\begin{center}
\begin{tabular}{|l|l|c|c|c|c|c|}
\hline
\textbf{Dataset} & \textbf{Method} & \textbf{Spleen} & \textbf{Kidney L} & \textbf{Kidney R} & \textbf{Liver} & \textbf{mean} \\
\hline
\hline

\multirow{7}{*}{ABD-110} & PANet-init \cite{wang2019panet}          & 30.95$\pm$1.09           & 19.24$\pm$0.37             & 17.64$\pm$0.71             & 49.91$\pm$0.34          & 29.43         \\
& PANet \cite{wang2019panet}            & 35.89$\pm$1.75           & 40.22$\pm$1.71             & 41.54$\pm$0.82             & 52.36$\pm$0.60          & 42.50         \\
& SE-Net \cite{roy2020squeeze} & 29.48$\pm$1.07           & 37.48$\pm$2.08             & 37.53$\pm$1.97             & 19.09$\pm$0.36          & 30.89         \\
& SSL-ALPNet  \cite{ouyang2020self}       & 64.90$\pm$1.62           & 61.58$\pm$2.53             & 64.05$\pm$2.27             & 71.83$\pm$1.81          & 65.59         \\
& Affine  & 50.42$\pm$0.91 & 53.04$\pm$1.57 & 52.025$\pm$2.17 & 66.99$\pm$1.20 & 55.62 \\
& RP-Net (Ours)                & \textbf{78.77$\pm$0.64}           & \textbf{81.89$\pm$1.45}             & \textbf{85.12$\pm$0.98}             & \textbf{81.88$\pm$0.63}          & \textbf{81.91}         \\
\cline{2-7}
\hline
\hline

\multirow{6}{*}{ABD-30} & SE-Net \cite{roy2020squeeze}    & 0.23           & 32.83             & 14.34       & 0.27       & 11.91        \\
& PANet-init \cite{wang2019panet}        & 23.82           & 13.97       & 14.17    & 50.27     & 25.55       \\
& PANet \cite{wang2019panet}      & 25.59                     & 32.34             & 17.37           & 38.42  & 29.42         \\
& SSL-ALPNet \cite{ouyang2020self} & 60.25                     & 63.34             & 54.82          & 73.65   & 63.02         \\
& Affine             & 48.99$\pm$1.48           & 43.44$\pm$2.04      & 45.67$\pm$1.45   & 68.93$\pm$0.88     & 51.75         \\
& RP-Net (Ours)       & \textbf{69.85$\pm$2.34}   & \textbf{70.48$\pm$2.55}        & \textbf{70.00$\pm$0.89}    & \textbf{79.62$\pm$0.91}     & \textbf{72.48}       \\

& RP-Net (fixed support)       & 68.27           & 71.59        & 70.27    & 80.51     & 72.66       \\
\cline{2-7}
\hline
\hline
\multirow{6}{*}{ABD-MR} & SE-Net \cite{roy2020squeeze}       & 51.80                     & 62.11             & 61.32        & 27.43     & 50.66         \\
& PANet-init \cite{wang2019panet}         & 34.59           & 18.63        & 22.50    & 47.43     & 30.78       \\
& PANet \cite{wang2019panet} & 50.90                    & 53.45             & 38.64     & 42.26         & 46.33         \\
& SSL-ALPNet \cite{ouyang2020self}   & 67.02                 & 73.63             & 78.39   & 73.05           & 73.02         \\
& Affine                & 62.87$\pm$1.80                        & 64.70$\pm$4.71             & 69.10$\pm$1.15    & 65$\pm$1.65         & 65.41         \\

& RP-Net (Ours)                   & \textbf{76.35$\pm$0.66}                    & \textbf{81.40$\pm$2.10}             & \textbf{85.78$\pm$1.12}       & \textbf{73.51$\pm$1.55}      & \textbf{79.26}   \\    
& RP-Net (fixed support)       & 75.96           & 80.18        & 86.63    & 73.52     & 78.99       \\
\cline{2-7}
\hline
\end{tabular}
\end{center}

\caption{DSC comparison with other methods on ABD-110, ABD-30 and ABD-MR (unit: \%).}
\label{table:abd-110}
\end{table*}

{\small
\bibliographystyle{ieee_fullname}
\bibliography{egbib}
}